\documentclass[11pt]{article}

\usepackage{acl}

\usepackage{times}
\usepackage{latexsym}

\usepackage[T1]{fontenc}

\usepackage[utf8]{inputenc}
\usepackage{tikz}
\usepackage{microtype}

\usepackage{inconsolata}

\usepackage{graphicx}

\title{Exploring Parameter-Efficient Fine-Tuning and Backtranslation for the \\ WMT 25 General Translation Task}
\author{Felipe Ribeiro Fujita de Mello \\
  Ritsumeikan University, Osaka, Japan \\
  \texttt{is0596kh@ed.ritsumei.ac.jp} \\\And
  Hideyuki Takada \\
  Ritsumeikan University, Osaka, Japan \\
  \texttt{htakada@is.ritsumei.ac.jp} \\}

\author{
  \textbf{Felipe Ribeiro Fujita de Mello\textsuperscript{1}},
  \textbf{Hideyuki Takada\textsuperscript{1}}
\\
  \textsuperscript{1} Ritsumeikan University, Japan
\\
  \small{
    \textbf{Correspondence:} \href{mailto:is0596kh@is.ritsumei.ac.jp}{is0596kh@is.ritsumei.ac.jp}
  }
}

\begin{document}
\maketitle
\begin{abstract}
In this paper, we explore the effectiveness of combining fine-tuning and backtranslation on a small Japanese corpus for neural machine translation. Starting from a baseline English→Japanese model (COMET = 0.460), we first apply backtranslation (BT) using synthetic data generated from monolingual Japanese corpora, yielding a modest increase (COMET = 0.468). Next, we fine-tune (FT) the model on a genuine small parallel dataset drawn from diverse Japanese news and literary corpora, achieving a substantial jump to COMET = 0.589 when using Mistral 7B. Finally, we integrate both backtranslation and fine-tuning—first augmenting the small dataset with BT generated examples, then adapting via FT—which further boosts performance to COMET = 0.597. These results demonstrate that, even with limited training data, the synergistic use of backtranslation and targeted fine-tuning on Japanese corpora can significantly enhance translation quality, outperforming each technique in isolation. This approach offers a lightweight yet powerful strategy for improving low-resource language pairs.
\end{abstract}

\section{Introduction}

Neural MT for Japanese benefits from recent large language models (LLMs) and recipe-driven data augmentation, but publicly documented, \emph{small-corpus} workflows are scarce. This paper focuses on a minimalist, engineering-first pipeline that couples (i) supervised fine-tuning (FT) on a small Japanese corpus with (ii) backtranslation (BT) to expand coverage. Our objectives are: 
\begin{itemize}
    \item To give a \textbf{clear blueprint} that other researchers can adopt even with limited computing resources.
    \item To perform \textbf{transparent evaluation}, using well-established metrics such as \textbf{COMET} and \textbf{BLEU/chrF}.
\end{itemize}

\section{Related Work}

Research on improving neural machine translation (NMT) for Japanese has increasingly relied on two complementary techniques: backtranslation and fine-tuning. Early large-scale systems demonstrated that backtranslation is particularly effective for low-resource settings, as it leverages abundant monolingual corpora to generate synthetic parallel data. This method augments scarce bilingual datasets and helps reduce domain mismatch, which is a persistent challenge in English--Japanese translation.

\citet{kiyono-etal-2020-tohoku} investigated English--Japanese news translation at WMT 2020, showing that the combination of synthetic data through backtranslation and subsequent fine-tuning significantly improved performance over a baseline. Extending this line of work, \citet{le-etal-2021-illinois} explored fine-tuning with domain-specific corpora and demonstrated that backtranslation enhanced adaptation to the news domain in the WMT 2021 shared task. Their study highlighted the importance of tailoring fine-tuning schedules when working with Japanese corpora.

Further refinements were presented by \citet{morishita-etal-2022-nt5} in WMT 2022, who introduced a system that incorporated both extensive backtranslation and selective fine-tuning. Their approach confirmed that even moderate-scale synthetic corpora, when carefully integrated, yield measurable improvements in translation accuracy for Japanese. Similarly, \citet{kudo-etal-2023-skim} reported results from WMT 2023 where backtranslation and iterative fine-tuning were applied to robustly adapt transformer-based systems, demonstrating strong gains for English--Japanese translation.

In parallel, multilingual NMT research has also highlighted the value of backtranslation. \citet{xu-etal-2021-improving-multilingual} proposed an auxiliary language framework, leveraging backtranslation across multiple language pairs, including Japanese. Their results suggest that cross-lingual signals derived from backtranslation not only improve individual language directions but also enhance multilingual consistency.

These studies illustrate the central role of backtranslation in augmenting limited Japanese corpora and show that fine-tuning, when combined with synthetic data, can consistently raise translation quality. They provide the empirical foundation for our own work for low-resource Japanese NMT.

\section{System Architecture}

\begin{figure}[t]
  \centering
  \includegraphics[width=\columnwidth]{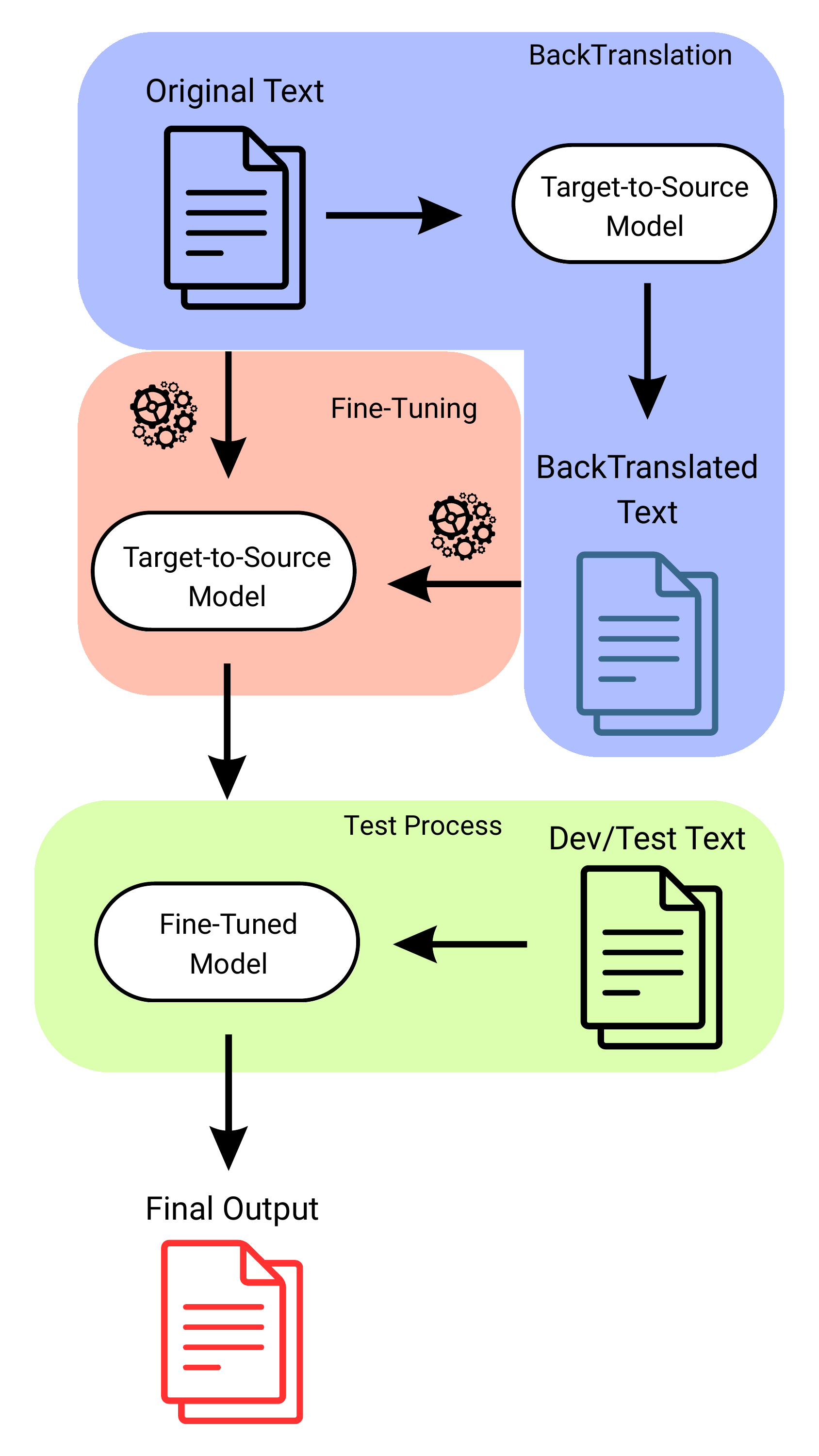}
  \caption{Overview of the proposed method}
  \label{fig:proposed-method}
\end{figure}

Our proposed method combines fine-tuning on a small parallel Japanese–English dataset with backtranslation to augment the available training data. As illustrated in Figure~\ref{fig:proposed-method}, monolingual Japanese sentences are first translated into English using the a pretrained model to create synthetic parallel pairs. These synthetic pairs are then used with the original data to fine-tune a pretrained model. The resulting system benefits from both the linguistic diversity of backtranslation and the domain adaptation of fine-tuning, leading to improved translation quality as measured by COMET, BLEU, and chrF++.

\subsection{Implementation Details}
The system builds on top of \texttt{AutoTokenizer} and \texttt{AutoModelForCausalLM}, enabling flexible experimentation with Mistral 7B\footnote{\url{https://huggingface.co/mistralai/Mistral-7B-v0.3}} \cite{jiang2023mistral7b}. Parameter-efficient fine-tuning is employed to reduce computational demands, while training routines follow established best practices with gradient accumulation, mixed precision (\texttt{torch.float16}), and GPU offloading.

\subsection{Dataset}
For our experiments, we relied on the Japanese--English \textit{WikiCorpus} released by Kyoto University\footnote{\url{https://alaginrc.nict.go.jp/WikiCorpus/}}. 
This corpus consists of parallel sentences extracted from Wikipedia, providing high-quality and naturally occurring examples of Japanese usage. 
Given the limited scope of our study, we sampled a total of approximately 1,500 sentence pairs for training and validation.

\subsection{Tokenization}
For Japanese text, we adopt \texttt{fugashi}\footnote{\url{https://github.com/polm/fugashi}}, a MeCab wrapper optimized for Python, which provides robust morphological analysis and segmentation. This ensures that the tokenizer can handle Japanese corpora effectively, producing consistent subword units that align with both training and backtranslation data.

\subsection{Backtranslation}

\begin{figure}[t]
  \centering
  \includegraphics[width=\columnwidth]{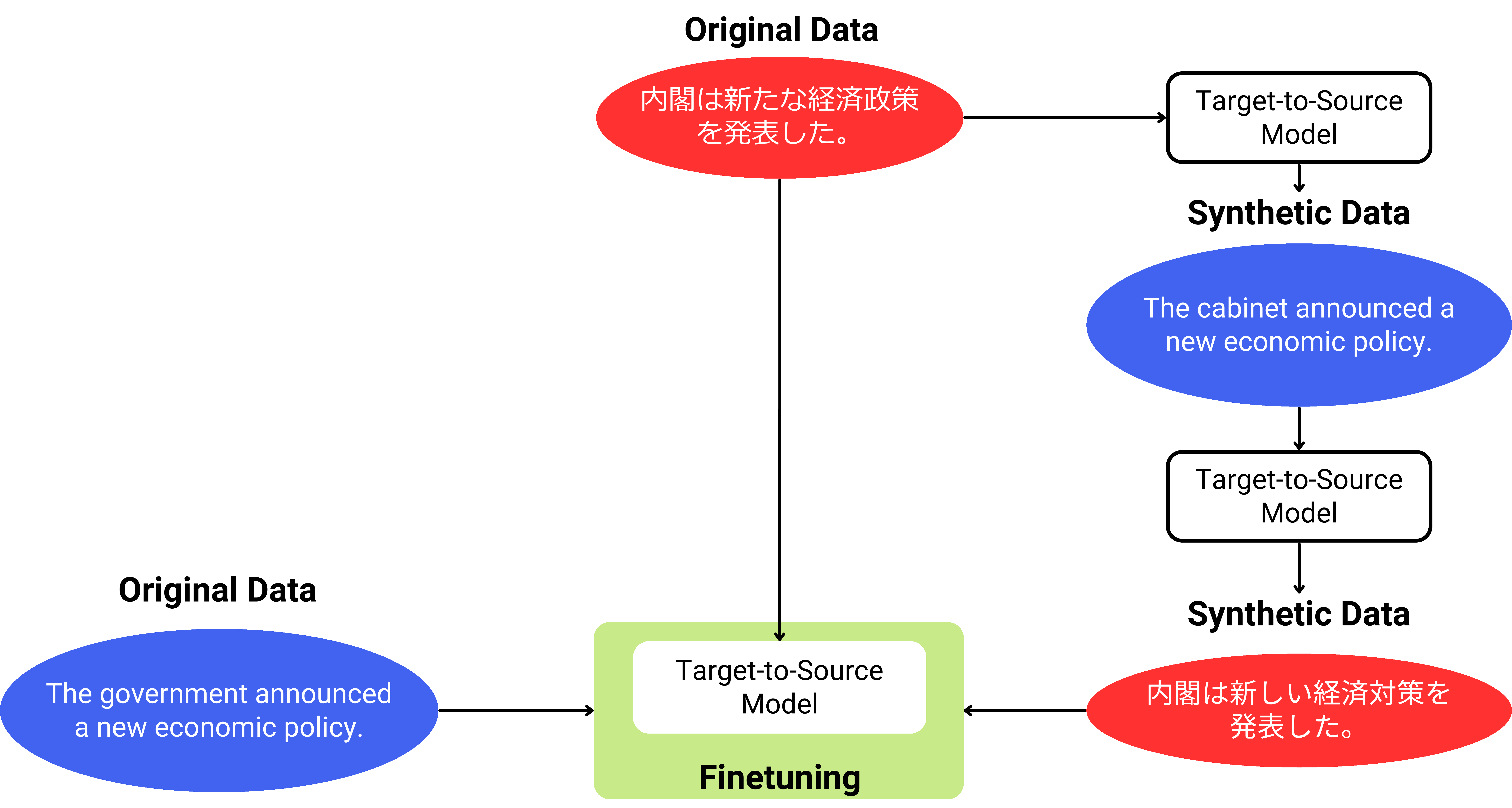}
  \caption{Overview of the backtranslation steps to generated synthetic data to serve as input along with the original data}
  \label{fig:bt}
\end{figure}

Backtranslation (BT) is implemented by first using a pretrained model (Japanese $\rightarrow$ English) on the available parallel data as shown in Figure~\ref{fig:bt}. Using this model, synthetic English sentences are generated from monolingual Japanese corpora. These synthetic pairs are then added to the original parallel dataset, effectively enlarging the training corpus. This augmentation proved crucial in mitigating data scarcity, providing additional coverage for domain-specific and colloquial expressions.

\subsection{Fine-Tuning Procedure}
Fine-tuning (FT) is performed on a small, high-quality parallel dataset of Japanese corpora. The fine-tuning focuses on adapting pre-trained Mistral 7B to the translation domain. To make this process more efficient, we employ parameter-efficient fine-tuning (PEFT) techniques, specifically Low-Rank Adaptation (LoRA) \cite{hu2021loralowrankadaptationlarge}. This approach enables effective adaptation to Japanese with limited resources, making fine-tuning feasible even under hardware constraints.

\subsection{Evaluation Metrics}
Evaluation is conducted using both automatic and human-oriented metrics. Automatic scores include:
\begin{itemize}
    \item \textbf{COMET} \cite{rei-etal-2020-comet}: a neural-based quality estimator, used as the primary evaluation metric. 
    \item \textbf{BLEU} \cite{papineni-etal-2002-bleu} and \textbf{chrF} \cite{popovic-2015-chrf}: reference-based metrics to provide comparability with prior work.
\end{itemize}
These metrics are computed on the validation set at each epoch and the final models are selected based on the best COMET score.

\subsection{System Configuration}
Our system is implemented using Hugging Face's \texttt{transformers} \footnote{\url{https://github.com/huggingface/transformers}} library. The key hyperparameters and settings are summarized in Table~\ref{tab:config}.
\begin{table}[h]
\centering
\small
\begin{tabular}{ll}
\hline
\textbf{Component}        & \textbf{Configuration} \\ \hline
Model                     & Mistral 7B (decoder-only) \\
Architecture              & Transformer decoder, 32 layers,\\
& hidden size 4096 \\
Training epochs           & 5--8 \\
Batch size                & 128 (with gradient accumulation) \\
Minibatch size            & 4 per device (before accumulation) \\
Learning rate             & $2 \times 10^{-5}$ (cosine schedule) \\
Max learning rate         & $3 \times 10^{-5}$ \\
Warmup steps              & 500 \\
Optimizer                 & AdamW \\
Weight decay              & 0.01 \\
Dropout                   & 0.1 \\
Gradient clipping         & 1.0 \\
Precision                 & Mixed (\texttt{float16}) \\
Decoding                  & Beam size 3, max new tokens 256, \\
                          & no sampling; length penalty 1.0 \\
Logging                   & Save best checkpoint on COMET \\
Number of updates         & 10,000 \\
\hline
\end{tabular}
\caption{System configuration for fine-tuning with backtranslation.}
\label{tab:config}
\end{table}

\section{Experiments}

\subsection{Setup}
We fine-tune on 1.5k seed pairs and their BT-augmented counterparts (same domain). We segment documents on blank lines, translate at paragraph level, and enforce paragraph-count parity. We then merge to document level, verify, and score.
Finally, we compared the results on several baselines to verify the system output compared to a state-of-art model.

\subsection{Results}

The results in Table~\ref{ftbt-table} show several consistent trends. First, applying backtranslation (BT) to the baseline Mistral 7B model provided only a marginal gain in COMET (0.468 vs. 0.460) while simultaneously lowering BLEU, suggesting that synthetic data alone cannot compensate for the absence of high-quality parallel supervision.

In contrast, fine-tuning (FT) on the small but high-quality Japanese parallel dataset yielded a substantial improvement, raising COMET to 0.589 and demonstrating the strong impact of targeted adaptation. When FT was combined with BT, the model achieved the highest COMET score of 0.597, confirming that the synergy between synthetic augmentation and fine-tuning is beneficial.

However, BLEU slightly decreased compared to FT alone, indicating that n-gram overlap metrics do not always align with adequacy-oriented metrics like COMET. This divergence highlights the importance of using multiple evaluation measures: while BLEU and chrF++ capture surface similarity, COMET better reflects semantic adequacy and fluency.

Overall, the results suggest that FT is the main driver of quality improvement in low-resource Japanese translation, while BT plays a supporting role by diversifying the training signal.

\begin{table}[ht]
\centering
\small
\begin{tabular}{lccc}
\hline
Model & BLEU & chrF & COMET \\
\hline
Mistral 7B Base        & 0.63 & --    & 0.460 \\
Mistral 7B Base + BT   & 0.18 & --    & 0.468 \\
Mistral 7B FT          & 1.97 & --    & 0.589 \\
Mistral 7B FT + BT     & 1.41 & 15.87 & 0.597 \\
\hline
\end{tabular}

\caption{Experimental results on Mistral 7B}
\label{ftbt-table}
\end{table}

\section{Limitations}

Our approach faces three main limitations. First, training on a small corpus makes the system highly sensitive to overfitting, requiring early stopping and regularization. Second, the effectiveness of backtranslation depends on the reverse model, as low-quality outputs can add noise; simple filtering methods such as length-ratio checks and language identification are necessary to maintain data quality. Finally, since the experiments rely on Wiki-derived text, there is a risk of domain shift when applying the model to other contexts, which may require domain adaptation.

\section{Conclusion}
In this work, we investigated the combined use of fine-tuning (FT) and backtranslation (BT) to improve English--Japanese neural machine translation under small-data conditions. The results show that parameter-efficient fine-tuning combined with carefully filtered backtranslation can provide a practical and effective blueprint for improving Japanese translation, even with limited computational resources. Future work will explore domain adaptation and scaling synthetic data generation to further enhance robustness.

\bibliography{custom}




\end{document}